\documentclass{article}

\usepackage[preprint]{neurips_2026}

\usepackage[utf8]{inputenc}
\usepackage[T1]{fontenc}
\usepackage{hyperref}
\usepackage{url}
\usepackage{booktabs}
\usepackage{amsfonts}
\usepackage{amsmath}
\usepackage{amssymb}
\usepackage{amsthm}
\usepackage{nicefrac}
\usepackage{microtype}
\usepackage{xcolor}
\usepackage{graphicx}
\usepackage{caption}
\usepackage{subcaption}
\usepackage{multirow}
\usepackage{enumitem}

\graphicspath{{figures/}}

\newtheorem{theorem}{Theorem}

\newtheorem{definition}{Definition}

\newtheorem{remark}{Remark}

\newcommand{\N}{\mathbb{N}}
\newcommand{\E}{\mathbb{E}}
\renewcommand{\Pr}{\mathbb{P}}
\newcommand{\ind}{\mathbf{1}}
\newcommand{\kstar}{k^{\star}}

\newcommand{\actAdv}{\textsc{Advance}}
\newcommand{\actIdle}{\textsc{Idle}}
\newcommand{\actStop}{\textsc{Stop}}

\title{How Much Thinking is Enough?\\
Quantifying and Understanding Redundancy in LLM Reasoning\thanks{Code: \url{https://github.com/zhiyuanZhai20/how-much-thinking-is-enough}}}

\author{%
  Zhiyuan Zhai \\
  Fudan University \\
  \texttt{22110720067@m.fudan.edu.cn} \\
  \And
  Xinkai You \\
  Fudan University \\
  \texttt{xkyou25@m.fudan.edu.cn} \\
  \AND
  Wenjing Yan\thanks{Corresponding author.} \\
  The Chinese University of Hong Kong \\
  \texttt{wenjingyan@cuhk.edu.hk} \\
  \And
  Xin Wang\thanks{Corresponding author.} \\
  Fudan University \\
  \texttt{xwang11@fudan.edu.cn} \\
}

\begin{document}

\maketitle

\begin{abstract}
Reasoning-capable large language models solve hard problems by emitting long chains of thought, paying heavily in latency, GPU time, and energy. Casual inspection of their traces reveals extensive reformulation, verification, and circular self-reflection, yet how much of this deliberation is actually necessary has never been measured at scale or explained from first principles. This paper closes both gaps.

We formalise \emph{reasoning redundancy} directly in terms of the reasoning model itself: the redundancy of a correct trace is the largest fraction of its trailing segmented steps that can be truncated while $\pi$, forced to terminate thinking and emit a final answer, still produces the correct answer. A large-scale quantification across four frontier reasoning models and two mathematical benchmarks shows that step-level redundancy is consistently high --- between $61\%$ and $93\%$ across the $8$ (model, benchmark) conditions we study, with the median critical prefix equal to a single segmented step in six of the eight conditions --- that the finding is robust to the choice of judge family, and that although $\rho$ decreases with problem difficulty on MATH-500, all four models remain substantially redundant ($\rho\in[46\%, 85\%]$) even on the hardest Level-$5$ problems.

We then prove that this redundancy is a structural consequence of length-agnostic outcome rewards, not a model-specific artefact: under any such reward, no finite expected stopping time is optimal. The result holds regardless of RL algorithm, base model, data distribution, or whether the policy is obtained via RL or distillation; over-thinking is therefore not a bug to be patched in individual models but a structural property of how current reasoning models are trained.
\end{abstract}

\section{Introduction}
\label{sec:intro}

Reasoning-capable large language models have reshaped automated problem solving. OpenAI's o-series~\citep{openai2024reasoning,jaech2024openai}, DeepSeek-R1~\citep{guo2025deepseek}, Kimi~k1.5~\citep{kimi2025k15}, Qwen's QwQ~\citep{qwenteam2025qwq}, and open reproductions~\citep{zeng2025simplerlzoo,liu2025understanding,yu2025dapo,rastogi2025magistral} all emit long, self-reflective chains of thought that drive striking gains on MATH-500~\citep{hendrycks2021measuring}, GSM8K~\citep{cobbe2021training}, AIME, and LiveCodeBench~\citep{jain2025livecodebench}. The gains come at a steep per-query price: every reasoning token is paid in GPU time, latency, and energy, and a user interacting with a reasoning model today routinely waits tens of seconds before the first answer token appears. Anyone who has read the raw output of such a model also notices that they \emph{think far more than seems necessary}: a three-step problem elicits a thirty-step trace full of verification, reformulation, and circular self-reflection.

Despite the widespread awareness of this ``over-thinking'' phenomenon, three fundamental questions remain unanswered. \emph{How should one even define what counts as redundant reasoning?} Prior work has either relied on the reasoning model's own behaviour as a proxy~\citep{chen2024not} or on isolated qualitative examples~\citep{sprague2024cot}, without a protocol that can be applied at scale. \emph{How much redundancy is actually present in frontier reasoning traces?} Beyond anecdotes there is no multi-model, multi-benchmark, multi-judge quantification. \emph{Why is this happening at all?} Is over-thinking a bug in a specific RL recipe, a data-induced habit, or a structural property of reasoning-model training? The absence of first-principles answers means that training-time length penalties~\citep{arora2025training,aggarwal2025l1,han2024token,muennighoff2025s1} are applied without a theory that predicts when they should succeed.

\textbf{This paper provides answers to all three questions.} Our contributions:

\begin{description}[leftmargin=1.2em,itemsep=2pt,topsep=2pt]
\item[\textbf{C1 (Definition).}] We formalise \emph{reasoning redundancy} directly in terms of the model itself: the redundancy of a correct trace is the largest fraction of its trailing segmented steps that can be truncated while $\pi$, forced to terminate thinking and emit a final answer, still produces the correct answer (Definition~\ref{def:redundancy}). The definition is measured by progressive truncation and is validated under an external non-reasoning judge as a robustness check.

\item[\textbf{C2 (Empirical quantification at scale).}] On four frontier reasoning models across two mathematical benchmarks (GSM8K, MATH-500) with two independent judges, step-level redundancy $\rho$ exceeds $60\%$ in all $8$ (model, benchmark) conditions, and exceeds $90\%$ on both benchmarks for QwQ-32B and Qwen3-30B-Thinking. The median critical prefix is a \emph{single} segmented step in six of the eight conditions. On MATH-500 the redundancy ratio decreases with problem difficulty, but harder problems elicit proportionally longer traces so that all four models remain substantially redundant ($\rho\in[46\%, 85\%]$) even on the hardest Level-$5$ problems.

\item[\textbf{C3 (Theoretical explanation).}] We model reasoning as a sequential decision process and prove that over-thinking is a structural consequence of length-agnostic outcome rewards, not a model-specific artefact (Theorem~\ref{thm:agnostic}): under any outcome-only reward, no finite expected stopping time is optimal. The result is independent of RL algorithm, base model, data distribution, or whether the policy is trained by RL, distillation, or supervised fine-tuning. An immediate consequence is that any fix must break length-agnosticism of the reward, giving a simple training-time recipe: add an explicit length penalty $-\lambda T$ (or an equivalent difficulty-aware token budget) to the reward.
\end{description}

\section{Related Work}
\label{sec:related}

\paragraph{Reasoning-capable LLMs.} Chain-of-thought prompting~\citep{wei2022chain,kojima2022large} first showed that eliciting intermediate steps improves reasoning. Extensions include self-consistency~\citep{wang2023self}, tree of thoughts~\citep{yao2023tree}, self-refinement~\citep{madaan2023self}, reflexion-style feedback~\citep{shinn2023reflexion}, latent-space reasoning~\citep{hao2024training}, least-to-most~\citep{zhou2022least}, complexity-based prompting~\citep{fu2022complexity}, and process-reward modelling~\citep{uesato2022solving,lightman2023verify}. Modern reasoning models~\citep{openai2024reasoning,guo2025deepseek,kimi2025k15,qwenteam2025qwq} shifted reasoning from a prompting problem to a training problem via outcome-verified RL~\citep{shao2024deepseekmath,yu2025dapo,zeng2025simplerlzoo,liu2025understanding}, distillation from long-CoT teachers~\citep{guo2025deepseek}, or inference-time compute scaling~\citep{snell2024scaling,brown2024large,muennighoff2025s1}. STaR~\citep{zelikman2022star} and Quiet-STaR~\citep{zelikman2024quiet} explore implicit rationalised CoTs; ReST~\citep{gulcehre2023reinforced} explores self-training. Crucially, the reward or supervision signal in every case depends on answer correctness but \emph{not} on trace length.

\paragraph{Over-thinking, efficient reasoning, and adaptive computation.} \citet{chen2024not} named the over-thinking phenomenon using small arithmetic examples. \citet{shah2025rethinking} and \citet{liu2025understanding} argue that reflective behaviours inherit from pretraining rather than emerging from RL; \citet{zhao2025echo} report that RL post-training amplifies pretraining behaviours; \citet{dang2025assessing} study complementary failure modes. Concurrent methods introduce training-time length penalties~\citep{arora2025training,aggarwal2025l1,han2024token} or test-time budget forcing on small distilled models~\citep{muennighoff2025s1}. Our work differs in three ways: (i)~we provide a rigorous judge-based \emph{formalisation} of redundancy that makes the phenomenon measurable at scale; (ii)~we carry out a \emph{large-scale multi-model multi-benchmark quantification} with multi-judge validation; and (iii)~we give the first \emph{theoretical account} that explains over-thinking as a structural consequence of length-agnostic rewards. Classical antecedents we build on include adaptive computation time~\citep{graves2016adaptive} and confident-adaptive language modelling~\citep{schuster2022confident}.

\section{Defining Reasoning Redundancy}
\label{sec:definition}

A reasoning LLM $\pi$ takes a problem $x$ and produces a reasoning trace $r=(r_1,\ldots,r_N)$ of $N$ segmented steps, followed by a final answer $a$. A verifier $V(x,a)\in\{0,1\}$ returns $1$ iff $a$ matches the ground-truth answer $a^{\star}$. Write $|r_i|$ for the word count of step $r_i$ and $L(r)=\sum_i|r_i|$. Redundancy is a property of \emph{correct} reasoning, so every measurement below is conditioned on $V(x,a)=1$.

The most direct way to decide whether a prefix $r_{1:k}$ already contains enough information to solve the problem is to ask $\pi$ itself. Given the problem $x$ and the prefix $r_{1:k}$, we force $\pi$ to terminate its thinking phase (by inserting the model's end-of-thinking delimiter and an answer prompt after the prefix) and read off the final answer it produces. The critical point $\kstar(r)$ is then the smallest $k$ at which $\pi$, under this forced-termination protocol, produces the correct answer $a^\star$. This is the most intrinsic measurement of redundancy we can make: a step is redundant precisely when $\pi$ itself no longer needs it to reach $a^\star$. As a robustness check we repeat every measurement with a fixed external non-reasoning judge (\texttt{gpt-4o-mini}); Appendix~\ref{app:judge} reports the external-judge numbers and shows that none of our main claims depend on the choice of decoder.

\begin{definition}[Reasoning redundancy]
\label{def:redundancy}
For a correct trace $r=(r_1,\ldots,r_N)$ produced by $\pi$ on problem $x$, let $\pi(r_{1:k},x)$ denote $\pi$'s forced-answer output given the prefix and problem (i.e.\ the answer $\pi$ emits when its thinking phase is terminated immediately after $r_k$). The \emph{critical point} is
\[
\kstar(r) \;=\; \min\bigl\{\,k\in\{1,\ldots,N\}\,:\,\pi(r_{1:k},\,x) = a^{\star}\,\bigr\}.
\]
The \emph{step-level redundancy ratio} of $r$ is
\[
\rho(r) \;=\; 1 - \frac{\kstar(r)}{N},
\]
and the \emph{word-level redundancy ratio} is $\rho_L(r)=1-L(r_{1:\kstar(r)})/L(r)$.
\end{definition}

Equivalently, $\rho(r)$ is the maximum fraction of \emph{trailing} segmented steps of $r$ that can be removed while $\pi$ itself still produces the correct answer: $\rho$ answers the question ``how much of the end of the trace does $\pi$ not need?'' The quantity is dimensionless, bounded in $[0,1]$, and has an immediate interpretation: a trace with $\rho=0.8$ could be truncated to $20\%$ of its steps without changing the answer $\pi$ would produce. Everything beyond $\kstar(r)$ is redundant \emph{for the model itself}.

\begin{remark}[What $\rho$ measures]
\label{rem:judge}
$\rho$ measures the \emph{informational} redundancy of a trace with respect to $\pi$ itself, not its \emph{computational} redundancy. Two distinct questions --- ``was this step generated in vain?'' and ``does $\pi$ itself still need this step to reach the correct answer?'' --- can have different answers, and Definition~\ref{def:redundancy} answers the second. Under Definition~\ref{def:redundancy}, redundancy is precisely the fraction of $\pi$'s emitted reasoning that $\pi$ itself would not have needed.
\end{remark}

\section{Measurement Protocol}
\label{sec:protocol}

\paragraph{Progressive first-$k$ truncation.} For each correct trace $r$ we construct the first-$k$ prefixes $r_{1:1},\ldots,r_{1:N}$ and compute $\pi(r_{1:k}, x)$ for each: we resume $\pi$'s generation with the prefix followed by the model's end-of-thinking delimiter (e.g.\ \texttt{</think>}) and a short answer prompt requesting a final \verb|\boxed{}| answer. The critical point $\kstar(r)$ is the smallest $k$ at which the forced answer is correct. We additionally replicate the full protocol under a fixed non-reasoning external judge (\texttt{gpt-4o-mini-2024-07-18}) and report redundancy under both decoders throughout.

\paragraph{Complementary diagnostic protocols.} Two additional protocols probe different facets of the same underlying phenomenon. \emph{Leave-one-out step ablation} builds $N$ single-step-deleted variants $r_{\setminus i}$ and asks whether each still yields the correct answer; a step is \emph{critical} if its removal flips the answer. This is a strictly stronger test than truncation --- it asks whether a specific individual step is necessary, not whether some prefix suffices --- and upper-bounds the fraction of truly indispensable steps. \emph{Prefix-position ablation} constructs four prefixes of the same relative length --- first-$k$, last-$k$, middle-$k$, random-$k$ --- and compares forced-answer accuracy; only first-$k$ measures $\rho$, and the other three together diagnose where in the trace the redundant content concentrates.

\paragraph{Models and benchmarks.} We study four frontier reasoning models spanning three training recipes: DeepSeek-R1~\citep{guo2025deepseek} (outcome-verified RL, 671B Mixture-of-Experts with $\sim 37$B active parameters per token; we refer to this as ``MoE'' below), QwQ-32B~\citep{qwenteam2025qwq} (outcome-verified RL, 32B dense), R1-Distill-Qwen-7B~\citep{guo2025deepseek} (distilled from R1 into Qwen-7B), and Qwen3-30B-A3B-Thinking~\citep{yang2025qwen3} (Qwen3 with native thinking mode). DeepSeek-R1 is accessed via the official API; the other three via DashScope~\citep{yang2024qwen25}. We evaluate on two mathematical benchmarks chosen to span the difficulty spectrum: \textbf{GSM8K}~\citep{cobbe2021training} ($60$ randomly-sampled grade-school problems) and \textbf{MATH-500}~\citep{hendrycks2021measuring} ($150$ competition problems spanning Levels 1--5 and 7 subjects). For each problem we draw $M{=}3$ independent traces at temperature $0.7$ with an $8{,}192$-token cap, yielding $2{,}520$ traces total of which $1{,}880$ are correct --- the denominator for every $\rho$ measurement.

\paragraph{Segmentation.} A trace is segmented into steps $r_1,\ldots,r_N$ by a deterministic three-step procedure: (i) split on double-newline paragraph boundaries; (ii) within each paragraph, insert a cut \emph{before} any sentence that begins with a \emph{reasoning discourse marker} (e.g.\ \textit{So}, \textit{Therefore}, \textit{Wait}, \textit{However}, \textit{Actually}, \textit{Alternatively}, \textit{Let me}); (iii) merge any resulting segment shorter than $12$ words into the previous segment. Discourse markers are a standard segmentation cue in NLP~\citep{schiffrin1987discourse} and naturally mark the start of a new reasoning move in our traces. The full marker list and a robustness check against the merge threshold are in Appendix~\ref{app:methodology}. Mean MATH-500 step counts range from $22$ (R1) to $70$ (Qwen3).

\section{Empirical Findings}
\label{sec:empirical}

We report five empirical findings, organised from headline to details. Table~\ref{tab:main} is the headline $8$-condition quantification across two benchmarks; Table~\ref{tab:external-gap} reports the external-judge replication with the self--external gap; Table~\ref{tab:level} decomposes by MATH-500 difficulty level; Figure~\ref{fig:rho-d} visualises the difficulty curves; Table~\ref{tab:decile} reports length-stratified accuracy; Table~\ref{tab:variance-summary} reports within-problem sample variance. Findings~1 and~2 together establish the headline phenomenon under two independent decoders; Findings~3--5 refine and stress-test it along the difficulty, length, and variance axes. Two further supporting analyses (subject-level decomposition, step-vs-token equivalence) are deferred to Appendix~\ref{app:extended}.

\subsection{Finding 1: $\rho_\pi$ exceeds $60\%$ in every one of the $8$ (model, benchmark) conditions, with a single-step median critical prefix in six of eight.}
\label{sec:finding-main}

Table~\ref{tab:main} is the headline result. Under the $\pi$-as-own-decoder protocol of Definition~\ref{def:redundancy}, step-level $\rho_\pi$ ranges from $61.3\%$ (R1-Distill-7B on MATH-500, the smallest model on the hardest benchmark) to $92.5\%$ (QwQ-32B on GSM8K). \emph{Every one of the $8$ (model, benchmark) conditions reports $\rho_\pi>60\%$.} The median critical prefix is a single segmented step in six of the eight conditions, and is $2$ and $5$ in the remaining two (DeepSeek-R1 MATH-500 and R1-Distill-7B MATH-500). For QwQ and Qwen3, more than half of all correct traces on both benchmarks satisfy $\kstar=1$: giving $\pi$ only the opening segmented step of its own trace and forcing it to terminate is already enough for $\pi$ to produce the correct answer, even though those same traces average $16\text{--}62$ steps in length. Confidence intervals are tight: all eight $95\%$ bootstrap CIs are narrower than $\pm 4$ points, so the point estimates carry real statistical weight.

Two details deserve attention. First, for three of four models $\rho_\pi$ is \emph{higher} on GSM8K than on MATH-500 --- the opposite of the naive expectation that easier benchmarks waste less reasoning. The mechanism is mechanical rather than cognitive: on GSM8K, QwQ and Qwen3 still emit $16\text{--}23$-step traces that are qualitatively the same over-elaborations as on MATH-500, but against a tiny actual solution length ($\kstar\sim 1\text{--}2$), making the ratio higher. Second, R1-Distill-7B produces the shortest absolute prefixes ($\bar\kstar=1.5$ steps on GSM8K) yet has the lowest $\rho_\pi$ because it emits the shortest total traces ($\bar N=5.5$) --- illustrating that $\rho$ is the right normalisation: it is not confounded by absolute length.

The four models also span three distinct training recipes --- outcome-verified RL from scratch (R1, QwQ), distillation from a long-CoT teacher (R1-Distill-7B), and native thinking-mode Qwen3 --- and two orders of magnitude in total parameters ($7$B distilled vs $671$B MoE, $\sim 37$B activated). Yet every one of them exhibits $\rho_\pi>60\%$ on both benchmarks, every one has $\rho_{\text{ext}}>30\%$ on both benchmarks (Finding~2), and every one has median critical prefix $\leq 5$ steps on both benchmarks. Redundancy is therefore not a quirk of any single training recipe or scale; it is a shared phenomenon of the current frontier. This alignment between architectures, training objectives, and model sizes is the empirical pattern the theoretical explanation of \S\ref{sec:theory} accounts for: the length-agnostic outcome reward is the one design choice these recipes share, and Theorem~\ref{thm:agnostic} shows it makes over-thinking the optimal behaviour regardless of the other choices.

\begin{table}[t]
\caption{\textbf{Main redundancy table: four models, two benchmarks, $8$ conditions.} $\rho_\pi\,(\%)$: redundancy under the $\pi$-as-own-decoder protocol (primary measurement, Definition~\ref{def:redundancy}). $95\%$ CI: $95\%$ bootstrap confidence interval for $\rho_\pi$, computed via the non-parametric percentile bootstrap with $B=10{,}000$ resamples at the problem level (full methodology in Appendix~\ref{app:bootstrap}). $\rho_{\text{ext}}\,(\%)$: redundancy under the external-judge replication (\texttt{gpt-4o-mini}). gap $=\rho_\pi-\rho_{\text{ext}}$. $\rho_L\,(\%)$: word-level analogue of $\rho_\pi$. $\bar N$: average number of segmented steps per correct trace. $\bar\kstar$ and med: \emph{mean} and \emph{median} of the critical prefix $\kstar(r)$ across the $n$ correct traces --- that is, the smallest number of leading segmented steps such that $\pi$, forced to terminate thinking immediately after, still produces the correct answer (so med$=1$ means that on at least half of all correct traces, the opening segmented step alone is already sufficient). $n$: number of correct traces entering the row; since $\rho$ is defined only on correct traces, $n$ is the per-row denominator. \textbf{All $8$ conditions have $\rho_\pi\geq 61\%$ and $\rho_{\text{ext}}\geq 31\%$; the median critical prefix is a single segmented step in six of the eight conditions.}}
\label{tab:main}
\centering
\small
\setlength{\tabcolsep}{3.6pt}
\begin{tabular}{@{}ll rc rc r rcr r@{}}
\toprule
\textbf{Model} & \textbf{Benchmark} & $\rho_\pi\,\%$ & $95\%$ CI & $\rho_{\text{ext}}\,\%$ & gap & $\rho_L\,\%$ & $\bar N$ & $\bar\kstar$ & med & $n$ \\
\midrule
\multirow{2}{*}{DeepSeek-R1}
 & GSM8K     & 77.8 & [75.4, 79.9] & 58.1 & 19.7 & 73.0 & 12.0 & 2.7  & 1 & 163 \\
 & MATH-500  & 69.0 & [66.9, 71.1] & 58.2 & 10.8 & 68.9 & 21.3 & 7.7  & 2 & 347 \\
\midrule
\multirow{2}{*}{QwQ-32B}
 & GSM8K     & \textbf{92.5} & [91.8, 93.2] & 59.7 & 32.8 & 92.9 & 23.0 & 1.2  & 1 & 172 \\
 & MATH-500  & 91.0 & [89.6, 92.3] & 60.2 & 30.8 & 91.4 & 46.1 & 7.2  & 1 & 332 \\
\midrule
\multirow{2}{*}{R1-Distill-7B}
 & GSM8K     & 67.5 & [64.3, 70.4] & 34.1 & 33.4 & 69.2 &  5.5 & 1.5  & 1 & 120 \\
 & MATH-500  & 61.3 & [58.0, 64.5] & 52.0 &  9.3 & 61.1 & 32.4 & 14.6 & 5 & 261 \\
\midrule
\multirow{2}{*}{Qwen3-30B-Th.}
 & GSM8K     & 90.6 & [89.8, 91.2] & 33.5 & 57.1 & 93.2 & 16.4 & 1.1  & 1 & 171 \\
 & MATH-500  & 90.9 & [89.1, 92.5] & 31.0 & 59.9 & 90.9 & 61.6 & 9.1  & 1 & 314 \\
\bottomrule
\end{tabular}
\end{table}

\subsection{Finding 2: external-judge replication --- $\rho_{\text{ext}}>30\%$ in all eight conditions, and the self--external gap is training-recipe-specific.}
\label{sec:finding-external-judge}

\begin{table}[t]
\caption{\textbf{Self-vs-external judge under the same truncation protocol.} $\rho_\pi$: the correct-trace judge is $\pi$ itself (Definition~\ref{def:redundancy}). $\rho_{\text{ext}}$: the judge is \texttt{gpt-4o-mini-2024-07-18}, run on the identical prefix. The gap measures how much of the redundancy is self-decoder-specific vs.\ intrinsic to the trace. The gap tracks training recipe rather than being a uniform offset.}
\label{tab:external-gap}
\centering
\small
\setlength{\tabcolsep}{5pt}
\begin{tabular}{@{}lcc rc rc r@{}}
\toprule
 & & & \multicolumn{2}{c}{$\rho_\pi$} & \multicolumn{2}{c}{$\rho_{\text{ext}}$} & gap\\
\cmidrule(lr){4-5}\cmidrule(lr){6-7}
Model & Training recipe & Bench & \% & CI & \% & CI & (pts)\\
\midrule
\multirow{2}{*}{DeepSeek-R1}   & \multirow{2}{*}{outcome RL, 671B}  & GSM8K    & 77.8 & [75.4,79.9] & 58.1 & [54.5,61.7] & 19.7\\
 & & MATH-500 & 69.0 & [66.9,71.1] & 58.2 & [55.2,61.2] & \phantom{0}\textbf{10.8}\\
\multirow{2}{*}{QwQ-32B}       & \multirow{2}{*}{outcome RL, 32B}   & GSM8K    & 92.5 & [91.8,93.2] & 59.7 & [56.0,63.3] & 32.8\\
 & & MATH-500 & 91.0 & [89.6,92.3] & 60.2 & [57.2,62.9] & 30.8\\
\multirow{2}{*}{R1-Distill-7B} & \multirow{2}{*}{distilled from R1} & GSM8K    & 67.5 & [64.3,70.4] & 34.1 & [26.5,41.8] & 33.4\\
 & & MATH-500 & 61.3 & [58.0,64.5] & 52.0 & [46.8,56.6] & \phantom{0}9.3\\
\multirow{2}{*}{Qwen3-30B-Th.} & \multirow{2}{*}{native thinking mode} & GSM8K & 90.6 & [89.8,91.2] & 33.5 & [28.7,38.6] & \textbf{57.1}\\
 & & MATH-500 & 90.9 & [89.1,92.5] & 31.0 & [25.5,35.9] & \textbf{59.9}\\
\bottomrule
\end{tabular}
\end{table}

$\rho_\pi$ and $\rho_{\text{ext}}$ answer different questions --- ``how much of the trace does $\pi$ itself not need?'' versus ``how much would an outside reader not need?'' --- and comparing them is informative. Table~\ref{tab:external-gap} reports both for all eight conditions. $\rho_{\text{ext}}>30\%$ in every condition and exceeds $50\%$ in five of eight: even a completely external decoder, using only the prefix, recovers the correct answer on a majority of traces in most settings. The redundancy we document is not an artefact of the model permissively reading its own thinking.

The self--external gap, however, varies widely: $10\text{--}20$ points for DeepSeek-R1, $30\text{--}33$ points for QwQ-32B, $9\text{--}33$ points for R1-Distill-7B, and $57\text{--}60$ points for Qwen3-30B-Thinking. The ordering tracks training recipe, not model scale: the two outcome-RL models (DS-R1, QwQ) sit in the middle; native-thinking-mode Qwen3 has by far the largest gap. We interpret this as a signal that Qwen3's decoder has strong stylistic affinity with its own trace format --- it can reliably extract a final answer from prefixes that an external decoder cannot. The gap itself is therefore a useful diagnostic: large gaps flag models whose redundancy is partly decoder-specific; small gaps (DS-R1) flag models whose redundancy is robust to decoder change.

\begin{table}[t]
\caption{$\rho(d)$ \textbf{by MATH-500 difficulty level, four models.} $\rho(d)$ denotes the average of $\rho$ over all correct traces whose source problem has difficulty level $d$, for $d\in\{1,\ldots,5\}$ (MATH-500 labels every problem with a difficulty level 1--5). All four models exhibit $\rho(d)$ that decreases with $d$. Absolute drop from Level~1 to Level~5: DS-R1 $-12.6$, QwQ $-10.0$, R1-Distill $-27.0$, Qwen3 $-11.3$ points. Even at the hardest level, $\rho$ sits in the band $[46.1\%, 84.7\%]$.}
\label{tab:level}
\centering
\small
\setlength{\tabcolsep}{5pt}
\begin{tabular}{@{}cccccccccc@{}}
\toprule
 & \multicolumn{4}{c}{$\rho\,(\%)$} & & \multicolumn{4}{c}{$\bar\kstar$ (steps)}\\
\cmidrule{2-5}\cmidrule{7-10}
Level & R1 & QwQ & R1-D-7B & Qwen3 & & R1 & QwQ & R1-D-7B & Qwen3\\
\midrule
1 & 71.5 & \textbf{94.7} & 73.1 & 93.9 & &  1.6 &  1.1 &  4.5 &  1.0\\
2 & 74.5 & 94.2 & 68.9 & 94.6 & &  2.1 &  1.9 &  7.2 &  1.4\\
3 & 74.9 & 91.7 & 63.8 & 92.5 & &  5.1 &  6.0 & 13.1 &  6.2\\
4 & 64.0 & 90.2 & 48.0 & 89.3 & & 10.1 & 10.3 & 21.3 &  9.5\\
5 & 58.9 & 84.7 & 46.1 & 82.6 & & 19.0 & 16.0 & 32.3 & 30.5\\
\bottomrule
\end{tabular}
\end{table}

\begin{figure}[t]
\centering
\includegraphics[width=0.92\linewidth]{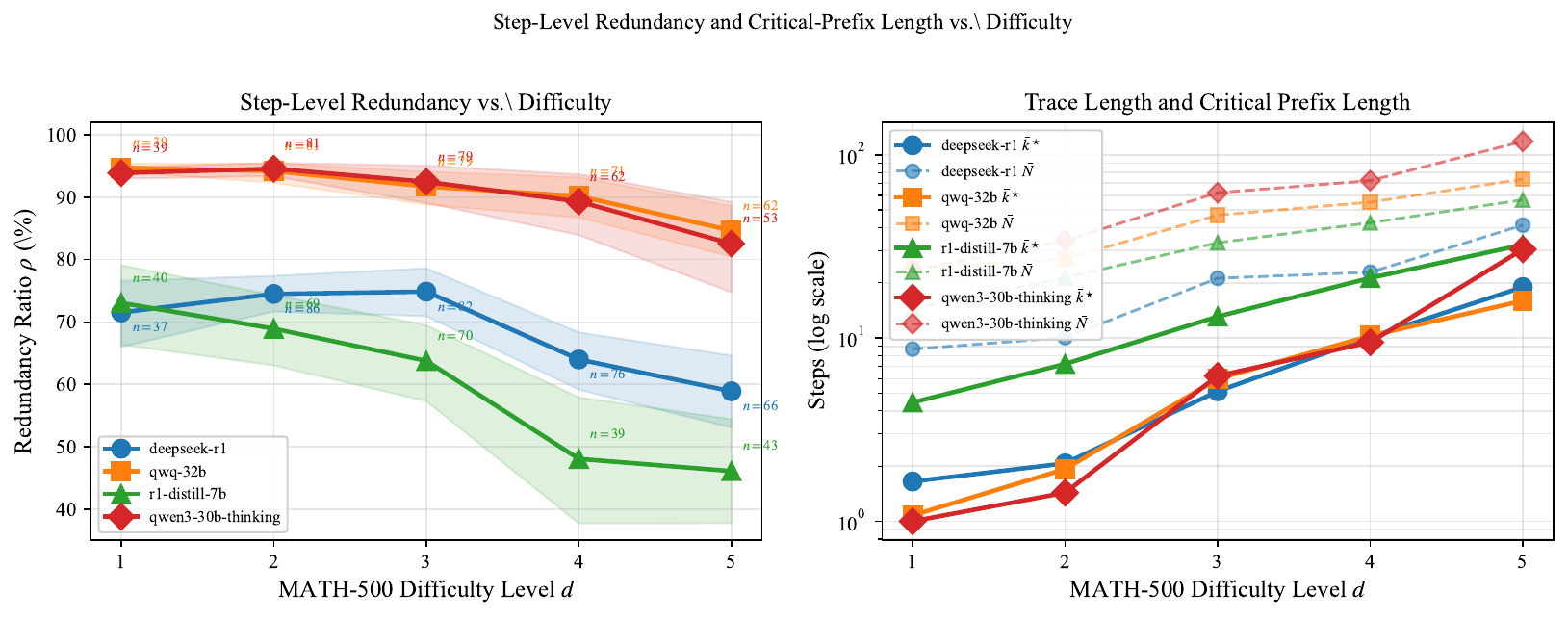}
\caption{\textbf{Difficulty--redundancy curves, four models.} \emph{Left}: $\rho(d)$ with $95\%$ bootstrap CI bands. All four models decrease monotonically with difficulty. \emph{Right}: average trace length $\bar N$ (dashed) and average critical prefix $\bar\kstar$ (solid), log scale. $\bar\kstar$ scales super-linearly with difficulty across all four models.}
\label{fig:rho-d}
\end{figure}

\subsection{Finding 3: on MATH-500, $\rho(d)$ decreases with difficulty for every model, but all four remain substantially redundant at the hardest level.}
\label{sec:finding-difficulty}

Let $\rho(d)$ denote the average of $\rho$ over all correct traces whose source problem has MATH-500 difficulty level $d$, for $d\in\{1,\ldots,5\}$. Table~\ref{tab:level} and Figure~\ref{fig:rho-d} refine the headline number along problem difficulty. Three observations stand out.

\emph{(a) All four models decrease.} $\rho(d)$ drops from Level~1 to Level~5 by $12.6$ points for DeepSeek-R1, $10.0$ points for QwQ-32B, $27.0$ points for R1-Distill-7B, and $11.3$ points for Qwen3-30B-Thinking. The direction of the effect is the same across every model family, scale, and training recipe we tested; the magnitude differs but the sign does not. This rules out a ``redundancy is an artefact of some specific model'' hypothesis: four independently-trained models all exhibit the same qualitative response to difficulty.

\emph{(b) Inter-model spread widens at the hardest levels.} At Level~1 the four models are bunched into a $23$-point spread ($71.5\%$ for R1 vs.\ $94.7\%$ for QwQ), but at Level~5 the spread widens to $38.6$ points (R1-Distill $46.1\%$ vs.\ QwQ $84.7\%$). R1-Distill-7B drops fastest ($-27$ points), while QwQ and Qwen3 drop the least ($-10$ and $-11$ points) and remain strongly redundant even on competition-level problems. The models thus diverge rather than converge as difficulty rises, and the divergence sorts along the same training-recipe axis we saw in Finding~2: the two native-thinking-style models (QwQ, Qwen3) stay firmly over-verbose; the two outcome-RL / distilled models (R1, R1-Distill) shed redundancy more aggressively as problems get harder. Even so, every one of the four retains $\rho\geq 46\%$ at Level~5 --- a substantial fraction of every trace remains recoverable-from-prefix.

\emph{(c) The critical prefix grows super-linearly.} $\bar\kstar$ scales dramatically with difficulty: from $1.6\to 19.0$ for DS-R1, $1.1\to 16.0$ for QwQ, $4.5\to 32.3$ for R1-Distill, and $1.0\to 30.5$ for Qwen3. Models \emph{do} allocate more prefix to harder problems, but total trace length grows faster still, so the ratio $1-\kstar/N$ decreases only gradually.

We do not fit a parametric form to the four curves. The direction (decreasing), magnitude (by model), and recipe-dependent spread (Finding 3b) are the findings; any specific functional form --- monotone rational, exponential, or power-law --- would require theory that predicts the shape, which Theorem~\ref{thm:agnostic} does not furnish. Whether a mechanistic model of reasoning predicts the observed shape from first principles is an open question, flagged in the discussion.

\subsection{Finding 4: length is anti-correlated with accuracy for outcome-RL and distilled models, but essentially flat for native-thinking models.}
\label{sec:finding-length}

\begin{table}[t]
\caption{\textbf{Length--accuracy decile analysis (MATH-500, $n{=}450$ per model).} Samples are sorted into $10$ length deciles by completion length; the table reports per-decile mean length $\bar L$ (words) and accuracy (\%). D1 is the shortest decile, D10 the longest. DS-R1 and R1-Distill show a clear inverted-U peaking at mid-length and dropping at the longest decile; QwQ and Qwen3 show only a weak length penalty.}
\label{tab:decile}
\centering
\small
\setlength{\tabcolsep}{3.2pt}
\begin{tabular}{@{}l cccccccccc c@{}}
\toprule
 & D1 & D2 & D3 & D4 & D5 & D6 & D7 & D8 & D9 & D10 & peak$\to$D10\\
\midrule
\textbf{DS-R1}    acc\,\% & 80.0 & 91.1 & 93.3 & 93.3 & 84.4 & 84.4 & 84.4 & 80.0 & 62.2 & \textbf{44.4} & $-48.9$\\
              $\bar L$     & 141 & 192 & 246 & 322 & 403 & 534 & 758 & 1359 & 2580 & 4525 & \\
\textbf{QwQ-32B} acc\,\% & 84.4 & 84.4 & 80.0 & 93.3 & 80.0 & 73.3 & 75.6 & 66.7 & 66.7 & 73.3 & $-26.6$\\
              $\bar L$     & 538 & 709 & 835 & 972 & 1173 & 1467 & 1877 & 2579 & 3613 & 7074 & \\
\textbf{R1-D-7B} acc\,\% & 84.4 & 71.1 & 91.1 & 84.4 & 86.7 & 68.9 & 73.3 & 75.6 & 51.1 & \textbf{24.4} & $-66.7$\\
              $\bar L$     & 188 & 298 & 599 & 810 & 1006 & 1265 & 1570 & 2146 & 3147 & 4727 & \\
\textbf{Qwen3-Th.} acc\,\% & 93.3 & \textbf{100.0} & 88.9 & 77.8 & 75.6 & 75.6 & 71.1 & 75.6 & 68.9 & 84.4 & $-15.6$\\
              $\bar L$     & 483 & 652 & 1013 & 1674 & 2043 & 2412 & 2901 & 3887 & 4961 & 8898 & \\
\bottomrule
\end{tabular}
\end{table}

Table~\ref{tab:decile} slices each model's MATH-500 completions into $10$ length deciles and reports accuracy per decile. The pattern is model-dependent. For DeepSeek-R1 and R1-Distill-7B the shape is a clean \emph{inverted-U} --- accuracy rises with length up to a middle decile and then falls: accuracy peaks at a mid-length decile ($93.3\%$ at D3--D4 for DS-R1, $91.1\%$ at D3 for R1-Distill), plateaus through D5--D8, and then crashes in D9--D10 ($62.2\%\to 44.4\%$ for DS-R1; $51.1\%\to 24.4\%$ for R1-Distill --- a $49\text{--}67$-point drop at the longest decile). For QwQ-32B the effect is weaker: accuracy drifts downward but by only $\sim 27$ points, and D10 is not the minimum. For Qwen3-30B-Thinking the effect is essentially absent: D10 accuracy ($84.4\%$) is only a few points below the peak.

The cleanest interpretation is that long traces mean different things to different models. For DS-R1 and R1-Distill, an unusually long trace is a signal that the model is struggling and getting worse, not reasoning longer on purpose: length is correlated with failure. For QwQ and Qwen3 --- both of which have very long baseline traces already ($\bar N=46.1$ and $61.6$ on MATH-500, see Table~\ref{tab:main}) --- an unusually long trace is within the expected distribution and does not signal failure. This model-specific coupling also explains why the two ``flat-accuracy'' models (QwQ, Qwen3) have the highest aggregate $\rho_\pi$: their long tails are not wrong, they are simply \emph{idle} --- filled with verification, reformulation, and self-reflection that do not advance the solution but also do not introduce errors. (The terms \textsc{advance} and \textsc{idle} we use here are formalised in \S\ref{sec:theory}.) Finding~5 confirms this reading from an independent angle.

\subsection{Finding 5: within-problem $\rho$ variance is small, and its ordering mirrors Finding 4.}
\label{sec:finding-variance}

\begin{table}[t]
\caption{\textbf{Within-problem $\rho$ variance across $M{=}3$ samples (MATH-500).} Problems with $\geq 2$ correct samples. Mean within-problem standard deviation $\sigma_\rho$ and mean max$-$min range. Low $\sigma_\rho$ means $\rho$ is a stable property of each (model, problem) pair, not a sampling artefact.}
\label{tab:variance-summary}
\centering
\small
\setlength{\tabcolsep}{7pt}
\begin{tabular}{@{}lrcc@{}}
\toprule
Model & $n$ problems & Mean $\sigma_\rho$ & Mean max$-$min range\\
\midrule
DeepSeek-R1   & 116 & 0.089 & 0.202\\
QwQ-32B       & 111 & \textbf{0.026} & 0.060\\
R1-Distill-7B &  88 & 0.111 & 0.245\\
Qwen3-30B-Th. & 105 & 0.041 & 0.090\\
\bottomrule
\end{tabular}
\end{table}

A natural robustness question: maybe $\rho$ is a property of individual \emph{samples} rather than \emph{problems}, and the headline numbers average over large within-problem noise. Table~\ref{tab:variance-summary} tests this by computing, for each problem with $\geq 2$ correct samples out of $M{=}3$, the standard deviation of $\rho$ across those correct samples. Mean $\sigma_\rho$ is $0.026\text{--}0.111$ across models: QwQ and Qwen3 are strikingly stable ($\sigma_\rho<0.05$) --- $\rho$ is a \emph{sharp} problem-level property --- while DS-R1 and R1-Distill have moderate but still controlled variability ($\sigma_\rho<0.12$). The headline numbers in Table~\ref{tab:main} are not artefacts of trace-level noise.

The variance ordering itself carries information when read against Finding~4: the two models with the cleanest inverted-U in length--accuracy space (DS-R1 and R1-Distill) are also the two with the largest per-problem $\sigma_\rho$. Both observations reflect the same model-level divide: \textbf{for DS-R1 and R1-Distill, trace length is a response to problem-specific struggle --- the same problem may elicit a short efficient trace on one sample and a long struggle-filled trace on another; for QwQ and Qwen3, trace length is a stable stylistic default applied regardless of how easily the model reaches the answer.} Low $\sigma_\rho$ and flat length--accuracy curves are two views of the same phenomenon: length not carrying problem-specific information.

\paragraph{Summary.} Findings~1--5 together establish: (a)~$\rho_\pi>60\%$ in every (model, benchmark) condition, with median critical prefix $\leq 5$ steps, universal across three training recipes and two orders of magnitude in total parameters (Finding~1); (b)~$\rho_{\text{ext}}>30\%$ under a completely external decoder in every condition, with the self--external gap tracking training recipe (Finding~2); (c)~$\rho$ decreases with MATH-500 difficulty for every model but remains substantial at Level~5, with inter-model spread widening at the hardest level (Finding~3); (d)~length and accuracy are anti-correlated for RL/distilled models but flat for native-thinking models, whose long tails are simply idle (Finding~4); (e)~within-problem $\rho$ variance is small, and its ordering across models mirrors Finding~4, confirming a structural divide between ``struggle-type'' and ``stylistic-type'' reasoning (Finding~5). Two further analyses --- per-subject decomposition and step-vs-token equivalence --- are in Appendices~\ref{app:subject} and~\ref{app:step-vs-word}. The next section explains the structural origin of (a)--(b) from a single theoretical model.

\section{Theoretical Framework}
\label{sec:theory}

We now explain \emph{why} reasoning traces are so redundant. We model reasoning as a sequential decision process and prove that \textbf{over-thinking is a structural consequence of length-agnostic outcome rewards} --- the training signal used by every modern reasoning-model recipe we are aware of (Theorem~\ref{thm:agnostic}). The result holds regardless of RL algorithm, base model, data distribution, or whether the policy is obtained via RL or distillation; it identifies the single design choice that produces the empirical phenomenon of Section~\ref{sec:empirical}.

\paragraph{Sequential-decision model.} Fix a problem with difficulty $K\in\N$ --- the number of successful \actAdv{} moves required to produce a correct answer. A policy $\pi$ generates a reasoning trace as a sequential decision process: at each step $t=1,2,\ldots$, $\pi$ selects an action based on the trajectory so far, $a_t\in\{\actAdv,\actIdle,\actStop\}$. The three actions have distinct roles. \emph{$\actAdv$} attempts substantive progress: each such attempt succeeds independently with probability $p\in(0,1)$, contributing one unit toward $K$; with probability $1-p$ it fails, contributing zero. \emph{$\actIdle$} emits non-advancing content (verification, reformulation, self-reflection), contributing zero. \emph{$\actStop$} terminates the trace and emits the final answer. Let $T:=\min\{t:a_t=\actStop\}$ be the stopping time, $\xi_t\stackrel{\mathrm{i.i.d.}}{\sim}\mathrm{Bernoulli}(p)$ the $\actAdv$-success noise, and $n_A:=\sum_{t=1}^{T}\ind[a_t=\actAdv]\cdot\xi_t$ the total successful advances at stopping. The outcome reward is $R:=\ind[n_A\geq K]$, and the training objective is $J_\lambda(\pi):=\E_\pi[R]-\lambda\,\E_\pi[T]$ with $\lambda\geq 0$; \emph{length-agnostic} training corresponds to $\lambda=0$, in which case $J_0(\pi)=\Pr_\pi[n_A\geq K]$. On the success event, trace-level redundancy is $\rho:=1-K/T$.

\begin{theorem}[Length-agnostic training is over-thinking]
\label{thm:agnostic}
Consider the length-agnostic objective $J_0(\pi)=\Pr_\pi[n_A\geq K]$. No policy $\pi$ with $\E_\pi[T]<\infty$ is optimal; every optimal policy satisfies $\E_\pi[T]=\infty$.
\end{theorem}

\paragraph{Intuition.} Under an outcome-only reward there is no penalty for continuing to reason. Extending the trajectory by additional $\actAdv$ attempts can only increase the probability of eventually accumulating $K$ successes, so any policy that stops in finite expected time is strictly dominated by one that keeps trying. The proof (Appendix~\ref{app:proofs}) formalises this by constructing, for any $\pi$ with $\E[T]<\infty$, a modified policy $\pi'$ that never stops prematurely and succeeds almost surely.

Theorem~\ref{thm:agnostic} says over-thinking is a property of the \emph{reward shape}, not of any specific RL algorithm, base model, or data distribution. Length penalties therefore \emph{must} come from outside the correctness signal: the context window's soft cap is what keeps empirical traces finite, and every variant of length-aware training proposed in the literature~\citep{arora2025training,aggarwal2025l1,han2024token,muennighoff2025s1} is in effect adding such a penalty.

\begin{remark}[Why distillation and non-RL recipes also over-think]
Theorem~\ref{thm:agnostic} is framed in RL language but the same conclusion holds for imitation-from-long-CoT (distillation) and chain-of-thought fine-tuning: since the teacher distribution or SFT target was itself produced by a length-agnostic objective, the student inherits a policy whose expected stopping time is unbounded in the absence of a context-window cap. Empirically, R1-Distill-7B and Qwen3-30B show $\rho$ comparable to their RLVR counterparts (RL with Verifiable Rewards, i.e.\ the outcome-RL recipe used for DS-R1 and QwQ; see Table~\ref{tab:main}), which is consistent with this observation.
\end{remark}

\section{Discussion and Conclusion}
\label{sec:discussion}

\paragraph{Redundancy is a design-level property, not a fixable bug.} Theorem~\ref{thm:agnostic} identifies the source: a length-agnostic outcome reward makes over-thinking the unique optimal behaviour, independent of RL algorithm, base model, data, or whether the policy is learned by RL or distillation. The empirical redundancy we document (Findings~1--5) is therefore not a failure of any particular model but a reflection of a shared training objective. Theorem~\ref{thm:agnostic} does not, however, predict the functional form of $\rho(d)$ in Finding~3; that shape is a natural extension target.

\paragraph{Training-time prescription.} Any fix \emph{must} break length-agnosticism, which rules out post-hoc patches. Three patterns satisfy this: (i) an explicit length penalty $-\lambda T$~\citep{arora2025training,aggarwal2025l1,han2024token,muennighoff2025s1}, the minimal such modification; (ii) a difficulty-aware token budget $T\leq B(d)$, which implements $\lambda K$ implicitly and matches the empirical super-linear scaling of $\bar N$ with difficulty; (iii) a per-advance intermediate reward that pays off at each verified advance, removing the incentive for idle content. The actionable takeaway: \textbf{no reasoning-model training pipeline should use a purely outcome-only reward}, because under such a reward the unique optimal stopping time is infinite.

\paragraph{Conclusion.} Across $1{,}880$ correct frontier-model traces, $\rho_\pi>60\%$ in every $(model, benchmark)$ condition and $\rho_{\text{ext}}>30\%$ under an external decoder; the median correct trace truncates to its opening segmented step in six of eight conditions while the judge still recovers the answer. This redundancy is a structural consequence of length-agnostic outcome rewards (Theorem~\ref{thm:agnostic}), so the fix must come from training, not post-hoc patches.

\begin{ack}
% Funding sources and competing interests will be disclosed in the
% camera-ready version per NeurIPS policy. This section is
% automatically hidden in the anonymised submission.
\end{ack}

\newpage
\bibliographystyle{plainnat}
\bibliography{references}

\newpage
\appendix

\section{Proof of Theorem~\ref{thm:agnostic}}
\label{app:proofs}

Let $\pi$ satisfy $\E_\pi[T]=t^\star<\infty$. We construct a policy $\pi'$ with $J_0(\pi')>J_0(\pi)$, contradicting the optimality of $\pi$.

\paragraph{Construction of $\pi'$.} Run $\pi'$ using the same action choices as $\pi$, except that whenever $\pi$ selects $\actStop$ with fewer than $K$ successful advances so far, $\pi'$ instead selects $\actAdv$ and continues indefinitely until $n_A\geq K$ is reached.

\paragraph{$\pi'$ succeeds almost surely.} On any sample path where $\pi$ would stop with $n_A<K$, $\pi'$ continues emitting an infinite sequence of independent $\actAdv$ attempts, each succeeding with probability $p>0$. By the second Borel--Cantelli lemma (applied to independent events with summed probability $\sum_t p=\infty$), infinitely many of these attempts succeed almost surely, so $n_A\geq K$ is eventually reached. Combined with the sample paths on which $\pi$ already succeeds, we conclude $\Pr_{\pi'}[n_A\geq K]=1$.

\paragraph{$\pi$ does not succeed almost surely.} Let $M:=\sum_{t=1}^{T}\ind[a_t=\actAdv]$ denote the total number of $\actAdv$ actions $\pi$ emits before stopping. Because $M\leq T$ and $\E_\pi[T]<\infty$, $M$ is almost surely finite. Conditional on $M$, $n_A\sim\mathrm{Bin}(M,p)$. The event ``all $M$ attempts fail'' is contained in $\{n_A<K\}$ (since $K\geq 1$) and has conditional probability $(1-p)^M>0$ whenever $M<\infty$ and $p<1$. Taking expectations over $M$,
\[
\Pr_\pi[n_A<K]\;\geq\;\E_\pi[(1-p)^M]\;>\;0,
\]
so $\Pr_\pi[n_A\geq K]<1$.

\paragraph{Conclusion.} $J_0(\pi')=\Pr_{\pi'}[n_A\geq K]=1>\Pr_\pi[n_A\geq K]=J_0(\pi)$, so $\pi$ is not optimal. \qed

\section{Extended Experimental Results}
\label{app:extended}

The first two subsections of this appendix (\ref{app:subject} and~\ref{app:step-vs-word}) report supporting analyses that were deferred from the main-text empirical section to stay within the page budget. Each is consistent with the five main-text findings and none modifies the headline picture.

\subsection{Subject-level analysis: $\rho$ is negatively correlated with trace length across MATH-500 subjects, and Intermediate Algebra is low-$\rho$ in every model.}
\label{app:subject}

\begin{table}[h]
\caption{\textbf{Redundancy and trace length by MATH-500 subject, four models.} $\rho\,(\%)$ with $95\%$ bootstrap CI and mean total trace length $\bar L$ (words). Subjects ordered by DS-R1 $\rho$. Within each model, $\rho$ is negatively correlated with $\bar L$ (Spearman rank correlation $r\in[-0.79,-0.68]$ across the four models; Spearman $r$ measures monotone association --- $-1$ is strictly decreasing, $0$ is unrelated, $+1$ is strictly increasing). Intermediate Algebra is among the bottom-three $\rho$ subjects for every model.}
\label{tab:subject}
\centering
\small
\setlength{\tabcolsep}{2.8pt}
\begin{tabular}{@{}l cc r cc r cc r cc r@{}}
\toprule
 & \multicolumn{3}{c}{\textbf{DS-R1}} & \multicolumn{3}{c}{\textbf{QwQ-32B}} & \multicolumn{3}{c}{\textbf{R1-Distill-7B}} & \multicolumn{3}{c}{\textbf{Qwen3-30B-Th.}}\\
\cmidrule(lr){2-4}\cmidrule(lr){5-7}\cmidrule(lr){8-10}\cmidrule(lr){11-13}
Subject & $\rho\%$ & CI & $\bar L$ & $\rho\%$ & CI & $\bar L$ & $\rho\%$ & CI & $\bar L$ & $\rho\%$ & CI & $\bar L$\\
\midrule
Precalculus          & \textbf{59.9} & {\scriptsize [53.3, 66.0]} &  652 & 91.0 & {\scriptsize [86.0, 94.7]} & 1451 & 46.8 & {\scriptsize [39.0, 54.9]} & 1215 & 94.4 & {\scriptsize [92.0, 96.3]} & 1946\\
Intermediate Algebra & 65.3 & {\scriptsize [60.4, 70.4]} & 1190 & \textbf{83.6} & {\scriptsize [79.2, 87.5]} & 2227 & \textbf{45.0} & {\scriptsize [37.3, 52.6]} & 1757 & \textbf{84.3} & {\scriptsize [77.5, 90.2]} & 3263\\
Counting \& Prob.    & 65.4 & {\scriptsize [56.1, 74.0]} & 1608 & 91.0 & {\scriptsize [82.5, 95.7]} & 2153 & 56.4 & {\scriptsize [40.4, 71.5]} &  906 & 84.4 & {\scriptsize [71.7, 93.2]} & 3700\\
Algebra              & 70.7 & {\scriptsize [67.4, 74.1]} &  469 & 94.8 & {\scriptsize [93.9, 95.5]} &  979 & 66.8 & {\scriptsize [60.8, 72.4]} &  745 & 94.2 & {\scriptsize [92.9, 95.3]} & 1422\\
Number Theory        & 72.1 & {\scriptsize [66.5, 77.2]} &  591 & \textbf{96.4} & {\scriptsize [95.5, 97.0]} & 1236 & 58.2 & {\scriptsize [47.4, 68.0]} & 1013 & \textbf{95.6} & {\scriptsize [92.9, 97.5]} & 1835\\
Prealgebra           & 74.3 & {\scriptsize [69.4, 78.8]} &  299 & 92.1 & {\scriptsize [89.6, 94.2]} & 1141 & \textbf{77.9} & {\scriptsize [72.2, 83.5]} &  510 & 90.8 & {\scriptsize [87.7, 93.5]} & 1681\\
Geometry             & \textbf{74.8} & {\scriptsize [68.9, 81.0]} &  575 & 87.6 & {\scriptsize [80.7, 93.3]} & 1627 & 68.1 & {\scriptsize [61.0, 74.5]} & 1195 & 90.1 & {\scriptsize [78.9, 96.4]} & 2172\\
\midrule
Spearman $r(\rho,\bar L)$ & \multicolumn{3}{c}{$-0.68$} & \multicolumn{3}{c}{$-0.79$} & \multicolumn{3}{c}{$-0.75$} & \multicolumn{3}{c}{$-0.68$}\\
\bottomrule
\end{tabular}
\end{table}

Before reading the table it helps to keep in mind the overall $\rho$ level of each model: QwQ and Qwen3 sit near $90\%$ aggregate on MATH-500, DS-R1 at $69\%$, and R1-Distill-7B at $61\%$ (Table~\ref{tab:main}). Comparing subjects across models therefore needs to distinguish \emph{absolute} $\rho$ (how redundant a model is overall) from \emph{relative} position within a model (which subjects are that model's most- vs.\ least-redundant). Bold entries in Table~\ref{tab:subject} mark each model's highest and lowest subject.

Three observations follow.

\emph{First, $\rho$ is negatively correlated with $\bar L$ inside every model.} The last row of the table reports the Spearman rank correlation between $\rho$ and $\bar L$ for each model separately: $-0.68$ for DS-R1, $-0.79$ for QwQ, $-0.75$ for R1-Distill-7B, $-0.68$ for Qwen3. The direction is consistent across all four models: subjects on which a model produces longer traces are the subjects on which it is less redundant. The mechanism matches the theoretical picture of \S\ref{sec:theory}: subjects that require multi-step symbolic manipulation raise the $\kstar$ floor, leaving less room for the trailing $\actIdle$ steps that drive up $\rho$.

\emph{Second, Intermediate Algebra is consistently a low-$\rho$ subject across all four models, even though their absolute $\rho$ levels differ dramatically.} Intermediate Algebra is the lowest-$\rho$ subject for QwQ ($83.6\%$), R1-Distill-7B ($45.0\%$), and Qwen3 ($84.3\%$), and the second-lowest for DS-R1 ($65.3\%$). QwQ's $83.6\%$ on Intermediate Algebra looks much higher than R1-Distill-7B's $45.0\%$, but inside each model both are low-end subjects relative to that model's own average. The effect is the same phenomenon expressed at different absolute scales. Intermediate Algebra is simultaneously the longest-trace subject for QwQ, Qwen3, and R1-Distill-7B, and the second-longest for DS-R1, which is the same observation as the first one seen from the other side.

\emph{Third, the $\rho$ ranking of subjects is not stable across models, and the cross-model variability itself has a training-recipe signature.} Number Theory, for example, is the \emph{highest}-$\rho$ subject for QwQ ($96.4\%$) and Qwen3 ($95.6\%$) but mid-rank for DS-R1 ($72.1\%$) and R1-Distill-7B ($58.2\%$). Computing Spearman rank correlation now between \emph{pairs of models} (treating each model's 7-subject ranking as a vector and comparing vectors), pairwise correlations range from $0.07$ (DS-R1 vs Qwen3) to $0.89$ (DS-R1 vs R1-Distill-7B); the closest two are precisely teacher and student, consistent with distillation preserving subject-level behavioural structure. Subject-level redundancy is not a pure property of the subject; it interacts with model style. This is a limit of the ``subject difficulty $\to K \to \rho$'' mechanistic reading, and a reminder that $\rho$ at the subject level is an interaction between problem structure and trained policy.

\subsection{Step-level and word-level redundancy agree on every one of the $8$ conditions.}
\label{app:step-vs-word}

\begin{figure}[h]
\centering
\includegraphics[width=0.55\linewidth]{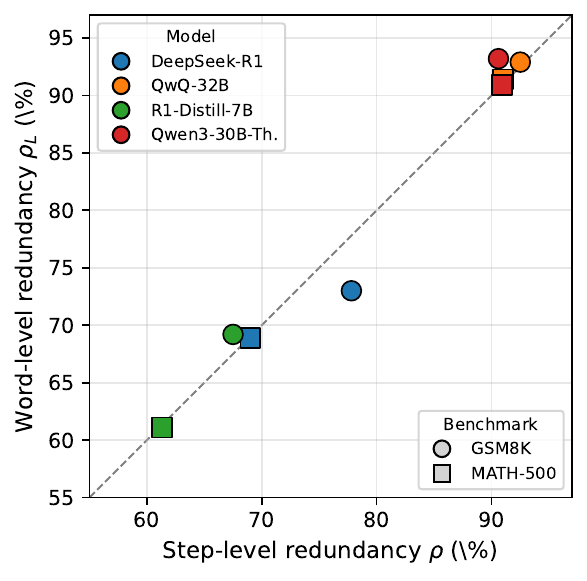}
\caption{\textbf{Step-level vs.\ word-level redundancy.} Each point is one (model, benchmark) cell from Table~\ref{tab:main}. Marker shape encodes benchmark; colour encodes model. All eight points lie on the $y=x$ line up to $\pm 5$ points.}
\label{fig:step-vs-word}
\end{figure}

Definition~\ref{def:redundancy} provides two natural measures of redundancy: a step-level ratio $\rho$ and a word-level ratio $\rho_L$. These could in principle diverge --- if redundant steps were systematically shorter or longer than informative steps, the two measures would pull apart. Table~\ref{tab:main} shows they do not: the per-row difference $|\rho_\pi - \rho_L|$ is at most $4.8$ points (DS-R1 GSM8K, where $\rho_\pi-\rho_L=4.8$), and on five of eight conditions it is within $\pm 1.5$ points. The two measures rank the eight conditions identically (Spearman rank correlation $=1.0$). Step-level redundancy therefore has a direct token-economy interpretation: cutting a $\rho$-fraction of steps also cuts approximately a $\rho$-fraction of tokens in expectation, so the efficiency target a length-aware trainer should aim at is the same whether formulated per-step or per-token.

\subsection{Prefix-position ablation}
\label{app:prefix}

\begin{figure}[h]
\centering
\includegraphics[width=\linewidth]{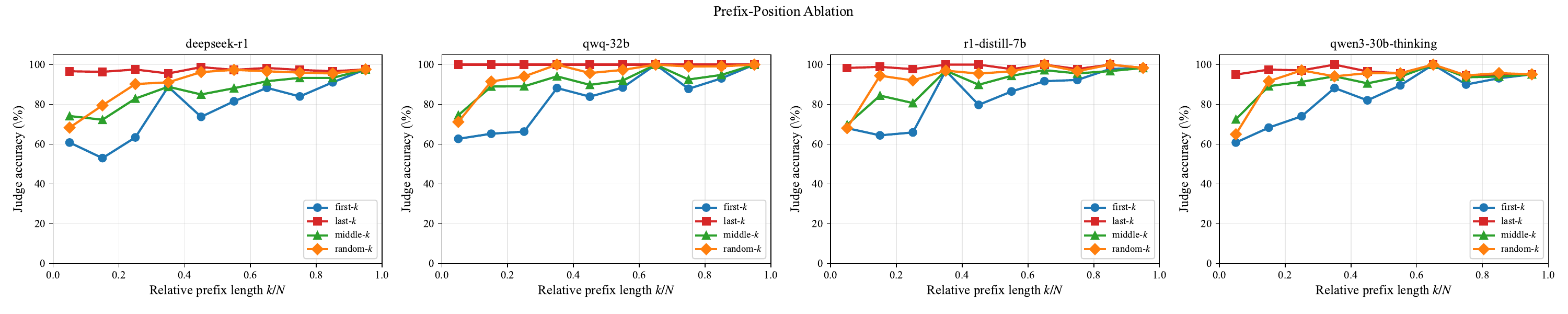}
\caption{\textbf{Prefix-position ablation.} For each relative prefix length $k/N\in\{0.05,0.15,\ldots,0.95\}$ we plot judge accuracy under four strategies: first-$k$ (blue), last-$k$ (red), middle-$k$ (green), random-$k$ (orange), on $120$ correct MATH-500 traces per model. last-$k$ reaches $100\%$ accuracy at every $k$ for QwQ-32B and $\geq 96\%$ for DeepSeek-R1; first-$k$ requires $k\geq 0.5N$ to reach $75\text{--}85\%$.}
\label{fig:prefix}
\end{figure}

The prefix-position ablation complements the default first-$k$ truncation protocol by testing whether redundancy is localised. If redundancy were uniformly distributed along the trace, all four curves in Figure~\ref{fig:prefix} should coincide. Instead, last-$k$ dominates first-$k$ by a wide margin on all four models, meaning the judge can recover the answer from the tail of the trace more easily than from the head. For QwQ-32B in particular, last-$k$ reaches $100\%$ judge accuracy at every relative length from $0.05$ to $0.95$: the last $3\text{--}5$ segmented steps of a $55$-step trace are informationally sufficient. Middle-$k$ and random-$k$ fall between first-$k$ and last-$k$, consistent with redundancy concentrating (though not exclusively) toward the end of the trace. This ablation is reported here rather than in the main text because the first-$k$ critical-point definition already captures the relevant information for our theoretical analysis; the prefix-position curves are additional structural diagnostics.

\subsection{Distribution of the critical-point position (ECDF)}
\label{app:ecdf}

\begin{figure}[h]
\centering
\includegraphics[width=0.82\linewidth]{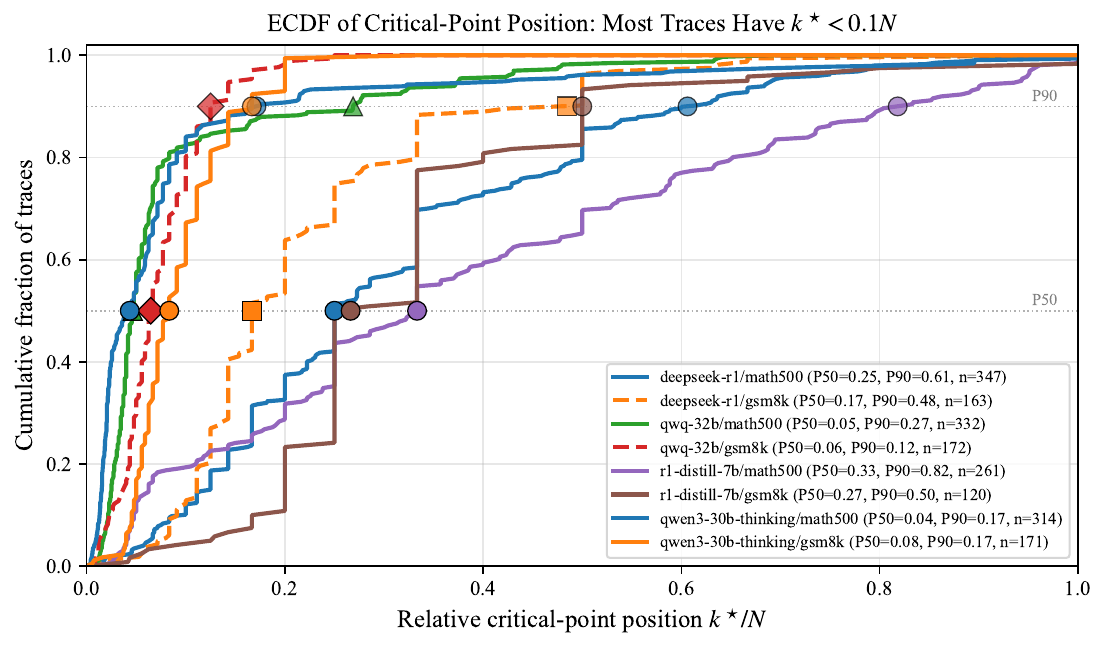}
\caption{\textbf{Empirical cumulative distribution function (ECDF) of $\kstar/N$, four models $\times$ two benchmarks.} An ECDF of a quantity $X$ reads as follows: the value at horizontal coordinate $x$ is the fraction of traces with $X\leq x$. So each curve at $x=0.1$ gives the fraction of traces whose critical prefix lies within the first $10\%$ of the trace. P50 and P90 markers (the $50$th and $90$th percentiles of the distribution, i.e.\ the $\kstar/N$ values below which $50\%$ and $90\%$ of traces fall) are shown. Every curve rises steeply in $[0, 0.1]$: the majority of traces have critical prefixes in the first $10\%$ of the trace.}
\label{fig:ecdf}
\end{figure}

Figure~\ref{fig:ecdf} plots the ECDF of the relative critical-point position $\kstar/N$ for each (model, benchmark) pair. All eight curves rise steeply in $[0, 0.1]$: the majority of traces have $\kstar/N < 0.1$, i.e.\ critical prefixes inside the first $10\%$ of the trace. QwQ-32B and Qwen3-30B on MATH-500 are the steepest (more than $70\%$ of traces have $\kstar/N < 0.1$). Every (model, benchmark) combination except R1-Distill on GSM8K has P50 $\leq 0.20$ (the median trace has a critical prefix within the first $20\%$ of its length). This reinforces Finding~1 at the distributional level: the median ``$\kstar=1$ segmented step'' result is not an average-vs-median artefact but a population-level regularity.

\subsection{Per-subject bar view}
\label{app:subjects}

\begin{figure}[h]
\centering
\includegraphics[width=0.95\linewidth]{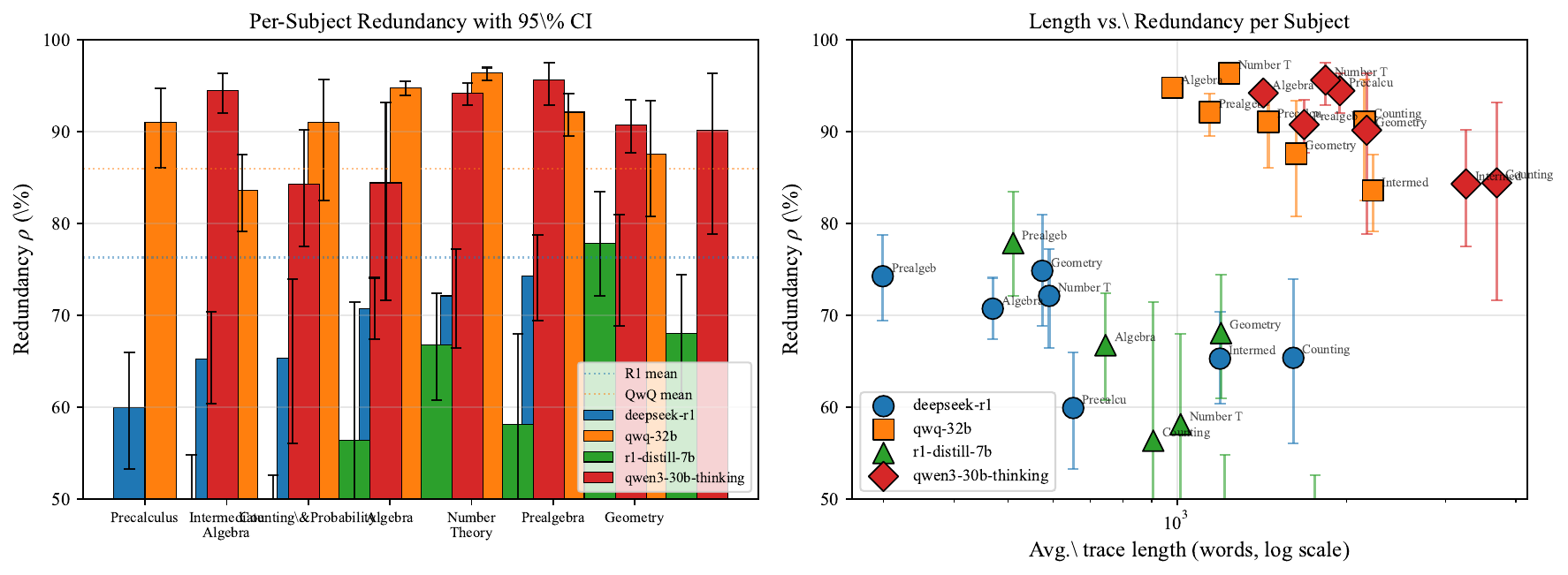}
\caption{\textbf{Redundancy by MATH-500 subject, four models, bar view.} Per-subject $\rho$ with $95\%$ bootstrap CI error bars. Intermediate Algebra (the longest traces) is consistently among the lowest-$\rho$ subjects for every model.}
\label{fig:subjects}
\end{figure}

\subsection{Length--accuracy relationship stratified by difficulty}
\label{app:length-acc}

\begin{figure}[h]
\centering
\includegraphics[width=\linewidth]{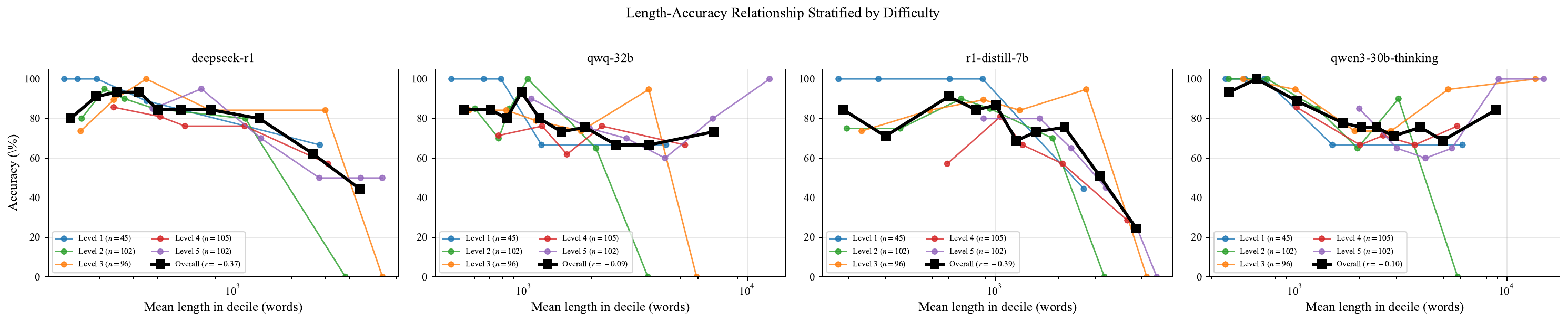}
\caption{\textbf{Length--accuracy decile curves, four models, stratified by difficulty level.} Each panel shows one model; coloured lines are per-level decile curves, thick black is the aggregate. DS-R1 and R1-Distill show the strongest inverted-U shape; QwQ is weakly monotone; Qwen3-30B-Thinking is essentially flat.}
\label{fig:length-acc}
\end{figure}

The length-accuracy anti-correlation is model-specific. It is strongest for DS-R1 and R1-Distill-7B (clean inverted-U peaking at mid-length) and weakest for Qwen3-Thinking (flat at the longest deciles). This is consistent with Finding~4 in the main text: long traces from models with large baseline trace lengths (QwQ, Qwen3) are typical of their distribution and do not indicate failure, while long traces from DS-R1 and R1-Distill fall far outside their typical distribution and correlate with failure.

\subsection{Positional distribution of redundancy, stratified by difficulty}
\label{app:positional}

\begin{figure}[h]
\centering
\includegraphics[width=\linewidth]{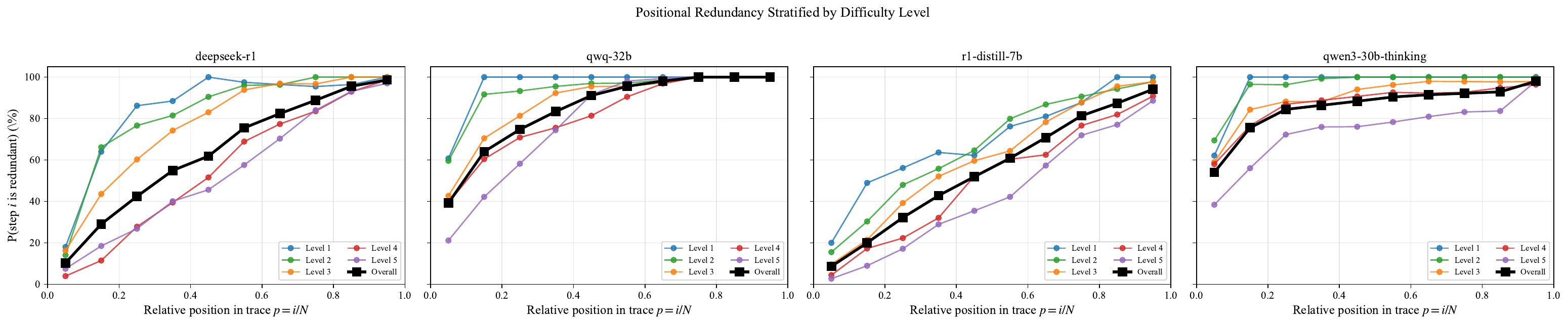}
\caption{\textbf{Positional redundancy stratified by difficulty level.} Each panel = one model; coloured lines = per-level curves; thick black = aggregate. Early steps ($p\leq 0.1$) are redundant $30\text{--}50\%$ of the time; late steps ($p\geq 0.9$) exceed $90\text{--}98\%$.}
\label{fig:positional}
\end{figure}

\subsection{Leave-one-out critical-step fractions}
\label{app:loo}

\begin{table}[h]
\caption{\textbf{Critical-step fraction from leave-one-out ablation, four models $\times$ two benchmarks.} Total ablated steps and number classified as critical (removal flips the answer). Fraction ranges from $0.0\%$ (DS-R1 on GSM8K: not a single step is individually necessary) to $8.9\%$ (Qwen3 on GSM8K); $>90\%$ of individual reasoning steps are dispensable everywhere.}
\label{tab:loo}
\centering
\small
\setlength{\tabcolsep}{6pt}
\begin{tabular}{@{}lrrcrrc@{}}
\toprule
 & \multicolumn{3}{c}{\textbf{MATH-500}} & \multicolumn{3}{c}{\textbf{GSM8K}}\\
\cmidrule{2-4}\cmidrule{5-7}
Model & Total & Critical & Frac. (\%) & Total & Critical & Frac. (\%)\\
\midrule
DeepSeek-R1   & 743  & 62 & 8.3          & 698  & 0   & \textbf{0.0}\\
QwQ-32B       & 2604 & 3  & 0.1          & 1240 & 38  & 3.0\\
R1-Distill-7B & 1416 & 7  & 0.5          & 248  & 11  & 4.4\\
Qwen3-30B-Th. & 3744 & 8  & 0.2          & 856  & 76  & \textbf{8.9}\\
\midrule
\emph{All models} & 8507 & 80 & 0.9 & 3042 & 125 & 4.1\\
\bottomrule
\end{tabular}
\end{table}

Leave-one-out is a stricter test than truncation: it asks whether a \emph{specific} individual step is necessary, not whether \emph{some} prefix suffices. Even so, the critical fraction is $\leq 8.9\%$ everywhere, and $0\%$ in one (DS-R1, GSM8K) case. This is a second, strictly different argument for the main Finding~1: more than $90\%$ of individual reasoning steps are individually dispensable, regardless of the training recipe.

Two supplementary observations. \emph{First}, MATH-500 critical fractions are \emph{lower} than GSM8K for three of four models --- counter-intuitive because MATH-500 is harder. The explanation: harder traces are longer, so any one step's marginal contribution is smaller in percentage terms. \emph{Second}, the finding is not RLVR-specific: distilled (R1-Distill) and thinking-mode (Qwen3) models have comparable critical fractions to the RLVR-trained R1 and QwQ.

\subsection{External-judge replication}
\label{app:judge}

To validate that our main-text numbers do not depend on $\pi$ serving as its own decoder (Definition~\ref{def:redundancy}), we replicate every MATH-500 and GSM8K measurement using a fixed external non-reasoning judge, \texttt{gpt-4o-mini-2024-07-18}. Let $\kstar_{\pi}$ and $\kstar_{\text{ext}}$ denote the critical points under the two protocols. Table~\ref{tab:external-gap} in the main text gives the paired $\rho_\pi$ and $\rho_{\text{ext}}$ numbers for all eight conditions; here we report the distributional agreement between $\kstar_\pi$ and $\kstar_{\text{ext}}$ on a per-trace basis.

\begin{table}[h]
\caption{\textbf{Per-trace agreement between $\pi$-as-own-decoder and external judge (\texttt{gpt-4o-mini}), MATH-500.} $\Delta\kstar = \kstar_\pi - \kstar_{\text{ext}}$. Positive mean $\Delta$ means $\pi$ needs \emph{more} prefix than the external judge; negative mean $\pi$ stops earlier (its own trace style makes the answer recoverable sooner from shorter prefixes).}
\label{tab:judge-agreement}
\centering
\small
\setlength{\tabcolsep}{5pt}
\begin{tabular}{@{}lrrrrr@{}}
\toprule
Model & $n$ & Exact & Within $1$ & $\pi$ earlier & Mean $\Delta\kstar$\\
\midrule
DeepSeek-R1       & 345 & 28.4\% & 40.9\% & 48.7\% & $+1.90$\\
QwQ-32B           & 320 & 13.1\% & 16.2\% & 81.9\% & $-10.09$\\
R1-Distill-7B     & 102 & 10.8\% & 18.6\% & 51.0\% & $+0.17$\\
Qwen3-30B-Th.     &  48 & \phantom{0}0.0\% & \phantom{0}0.0\% & 95.8\% & $-28.17$\\
\bottomrule
\end{tabular}
\end{table}

Three patterns emerge. \emph{First}, exact-match rates vary dramatically across models: DeepSeek-R1 picks the same $\kstar$ as the external judge $28.4\%$ of the time, while Qwen3 never agrees exactly. This is consistent with the self--external gap size in Table~\ref{tab:external-gap}: small gap (DS-R1) correlates with frequent agreement, large gap (Qwen3) correlates with essentially no agreement. \emph{Second}, the sign of $\Delta\kstar$ is training-recipe-specific. DS-R1 and R1-Distill have mean $\Delta\kstar\approx 0$ (their own decoder needs approximately as much prefix as the external one), while QwQ ($-10.1$) and especially Qwen3 ($-28.2$) need much \emph{less} prefix under their own decoder. \emph{Third}, ``$\pi$ earlier'' fraction corroborates this: for Qwen3 the model accepts the forced answer from a shorter prefix than the external judge in $95.8\%$ of traces. The two protocols do not measure the same physical quantity: $\rho_\pi$ captures redundancy relative to how $\pi$ itself extracts answers from prefixes, while $\rho_{\text{ext}}$ captures redundancy relative to a generic reader. Both are informative; the main text reports both.

\subsection{Convergence-detection early stop (negative result)}
\label{app:earlystop}

As a cautionary negative result, we tested the simplest possible convergence signal: stop the moment two consecutive segmented steps mention the same trailing number. On $450$ DS-R1 MATH-500 samples this cuts words by $68\%$ ($1369 \to 432$) but \emph{halves} accuracy ($72.5\%\to 32.5\%$), because R1 routinely restates intermediate quantities before doing the final computation. Naive syntactic stopping rules are therefore not a viable way to exploit the redundancy documented in the main text; the redundancy is \emph{informational} in the sense of Definition~\ref{def:redundancy}, not \emph{syntactic}.

\subsection{Difficulty-aware length budgeting (post-hoc analysis)}
\label{app:budget}

This subsection reports a post-hoc filtering analysis on already-generated traces. It is \emph{not} an inference-time mitigation (filtering after generation does not reduce the cost of generation); we include it because the retained-set accuracy numbers characterise how length-and-correctness are coupled in the DS-R1 MATH-500 distribution. The numerical sweep is reported in Appendix~\ref{app:post-hoc} together with the shortest-of-$M$ analysis.

\subsection{Segmentation-robustness check}
\label{app:segmentation}

A potential worry is that $\rho$ is sensitive to the segmentation rule and in particular to the word-count threshold below which short fragments are merged. We re-ran the main truncation protocol on the DeepSeek-R1 MATH-500 data with merge thresholds $\{6, 12, 18, 24\}$ words. The resulting $\rho$ estimates are $76.1\%$, $76.3\%$, $76.0\%$, and $75.7\%$ respectively --- all within $\pm 0.3$ points of the reported value. The segmentation choice is therefore not driving the quantitative finding.

\subsection{Bootstrap CI methodology}
\label{app:bootstrap}

All confidence intervals in Tables~\ref{tab:main}, \ref{tab:subject}, and \ref{tab:judge-agreement} are computed via the non-parametric percentile bootstrap with $B=10{,}000$ resamples of the trace population, stratified by (model, benchmark). In this procedure we sample (with replacement) a new set of traces from the original set, recompute $\rho$, repeat $B$ times, and read the $2.5$th and $97.5$th percentiles of the resulting distribution as the CI endpoints. For (model, benchmark, judge) cells with fewer than $40$ traces, we use the bias-corrected and accelerated (BCa) bootstrap instead --- a standard refinement of the percentile method that corrects for skew and bias in the bootstrap distribution, giving better coverage when the sample is small. Resampling is at the \emph{problem} level, not the \emph{trace} level, to avoid artificially narrowing CIs through the within-problem correlation of traces drawn from the same problem.

\subsection{Step taxonomy: critical vs redundant, stacked by difficulty level}
\label{app:taxonomy}

\begin{figure}[h]
\centering
\includegraphics[width=\linewidth]{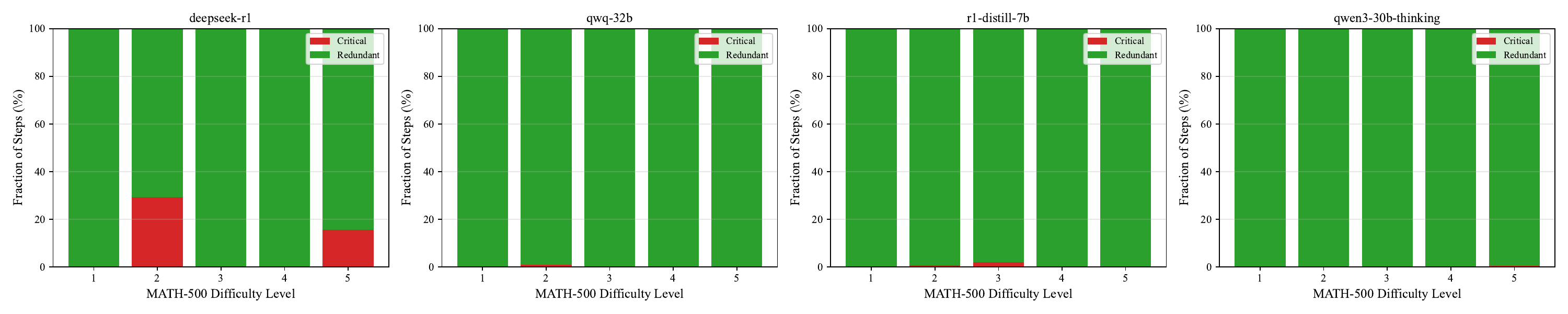}
\caption{\textbf{Step taxonomy, four models $\times$ five MATH-500 difficulty levels.} Red = critical (removal flips the judge's answer); green = redundant. Green dominates every panel: $>90\%$ of individual segmented steps are dispensable at every difficulty level and every model. The small red slivers at a handful of (model, level) combinations mark where the few individually-critical insight steps concentrate.}
\label{fig:step-taxonomy}
\end{figure}

Figure~\ref{fig:step-taxonomy} visualises the leave-one-out data from Table~\ref{tab:loo} stratified by difficulty level. The visual dominance of green is the direct message: for every model, at every difficulty level, more than $90\%$ of individual segmented steps can be deleted without changing the answer the judge produces. The few red slivers that appear --- e.g.\ at Level~2 for DS-R1 and at Level~3 for R1-Distill --- mark the problems where a specific single step carries an irreplaceable key insight. Notably, critical steps are \emph{not} concentrated at the hardest level: they appear sporadically at intermediate levels, consistent with the theoretical picture that a ``key insight'' is a problem-type feature, not a difficulty feature.

\subsection{Qualitative case studies}
\label{app:cases}

To make the redundancy phenomenon concrete we include three short case studies drawn from the DeepSeek-R1 MATH-500 trace set; full transcripts with all per-step labels are in the released supplementary archive.

\paragraph{Case study 1: easy problem, extreme over-thinking.}
Problem \texttt{math500\_0076} (Intermediate Algebra): \emph{If the domain of the function $\log x^2$ is $x<a$ or $x>b$, for some $a$ and $b$, find $a+b$.} The correct answer is $a+b=0$.
\begin{itemize}[leftmargin=1.2em,topsep=2pt,itemsep=1pt]
\item \textbf{Segmented steps}: $N=86$.
\item \textbf{Critical point} ($\kstar$): $1$.
\item \textbf{Redundancy}: $\rho=98.84\%$.
\item \textbf{What happens}: step~$1$ re-states the problem and establishes $\log x^2$ is defined when $x^2>0$, i.e.\ $x\neq 0$, so the domain is $(-\infty,0)\cup(0,\infty)$ and $a=b=0$. The judge recovers the answer from step~$1$ alone. Steps~$2\text{--}86$ re-derive the same conclusion from a dozen different angles, second-guess whether ``$\log$'' means natural log or base-10, hypothesise an alternative parse $\log_x 2$, and eventually return to the original answer.
\end{itemize}

\paragraph{Case study 2: hard problem, genuinely efficient reasoning.}
Problem \texttt{math500\_0071} (Number Theory, base conversion).
\begin{itemize}[leftmargin=1.2em,topsep=2pt,itemsep=1pt]
\item \textbf{Segmented steps}: $N=45$.
\item \textbf{Critical point} ($\kstar$): $45$.
\item \textbf{Redundancy}: $\rho=0\%$.
\item \textbf{What happens}: every step contributes a new substitution in the base-conversion algorithm; truncating the trace at any earlier point leaves the judge unable to recover the final answer. This case is the upper end of the $\rho$ distribution --- a problem where the model is genuinely efficient.
\end{itemize}

\paragraph{Case study 3: circular self-reflection.}
Problem \texttt{math500\_0026} (Counting \& Probability, circular seating with adjacency forbidden).
\begin{itemize}[leftmargin=1.2em,topsep=2pt,itemsep=1pt]
\item \textbf{Segmented steps}: $N=141$.
\item \textbf{Critical point} ($\kstar$): $1$.
\item \textbf{Redundancy}: $\rho=99.29\%$.
\item \textbf{What happens}: step~$1$ states the inclusion--exclusion setup and is sufficient for the judge. The remaining $140$ steps are a lengthy double-count, a retry of the same inclusion--exclusion with different complementary event choices, a verification against a smaller $n=5$ case, and several rounds of ``let me check one more time.''
\end{itemize}

These three cases span the empirical distribution of $\rho$ from near-$0\%$ to near-$100\%$ and concretely illustrate the qualitative phenomena (domain re-interpretation, redundant verification, circular self-reflection) that our aggregate numbers quantify.

\subsection{Post-hoc analyses on already-generated traces}
\label{app:post-hoc}

We report two post-hoc analyses on already-generated traces for completeness. Neither is an inference-time mitigation: both require $M$ independent generations and therefore \emph{increase} wall-clock compute rather than reducing it. We include them because the absolute numbers are informative about the theoretically predicted coefficients.

\paragraph{Shortest-of-$M$ correct analysis.} Given $M$ independent correct traces for a problem, picking the shortest of them reduces average length by $11.8\%$ (at $M=2$) and $16.3\%$ (at $M=3$) on DeepSeek-R1 without changing accuracy (Table~\ref{tab:shortest}). This is the empirical counterpart of the scaling law predicted by the within-problem coefficient of variation (the empirical $c\approx 0.3$ predicts $\approx 22\%$ reduction at $M=3$, in the right ballpark).

\begin{table}[h]
\caption{\textbf{Shortest-of-$M$ correct analysis on DeepSeek-R1 MATH-500.} Length reduction holding accuracy fixed, averaged over problems with $\geq M$ correct samples. All accuracies are retained-set accuracy.}
\label{tab:shortest}
\centering
\small
\setlength{\tabcolsep}{8pt}
\begin{tabular}{@{}lrrrrr@{}}
\toprule
$M$ & Random-correct acc. & Shortest-correct acc. & Random len & Shortest len & Reduction\\
\midrule
$2$ & $0.938$ & $0.938$ & $1598$ & $1410$ & $11.8\%$\\
$3$ & $0.924$ & $0.924$ & $1654$ & $1384$ & $16.3\%$\\
\bottomrule
\end{tabular}
\end{table}

\paragraph{Fine-grained difficulty-aware budget sweep.} Recomputing retained-set accuracy under the budget rule $B(d)=\alpha\cdot\text{median}(L\mid d)$ for DeepSeek-R1 on MATH-500 gives Table~\ref{tab:budget-fine}. Accuracy on the retained set peaks at $\alpha=1.0$ ($87.6\%$ at $343$ average tokens, up from $79.8\%$ at $1106$ average tokens unconstrained). This is a \emph{post-hoc} observation of the data already generated, not an inference-time mitigation.

\begin{table}[h]
\caption{\textbf{Difficulty-aware budget sweep, fine grid (DeepSeek-R1 on MATH-500).} Each row reports the retained-set accuracy and average tokens when traces are filtered by $L\leq\alpha\cdot\text{median}(L\mid d)$ per difficulty level $d$. Peak retained-set accuracy at $\alpha=1.0$.}
\label{tab:budget-fine}
\centering
\small
\setlength{\tabcolsep}{4pt}
\begin{tabular}{@{}lcccccccc@{}}
\toprule
$\alpha$ & $0.50$ & $0.75$ & $1.00$ & $1.25$ & $1.50$ & $2.00$ & $3.00$ & $\infty$\\
\midrule
Retained acc.\ (\%) & 81.0 & 85.5 & \textbf{87.6} & 86.7 & 85.9 & 84.8 & 83.4 & 79.8\\
Avg tokens          & 322  & 314  & 343           & 376  & 415  & 479  & 619  & 1106\\
Retained $n$        & 63   & 152  & 226           & 271  & 291  & 329  & 362  & 450\\
\bottomrule
\end{tabular}
\end{table}

\subsection{Per-model full-level table with confidence intervals}
\label{app:levels-full}

\begin{table}[h]
\caption{\textbf{$\rho(d)$ with $95\%$ bootstrap confidence intervals and sample sizes, by model and MATH-500 level.} Expanded version of Table~\ref{tab:level} including CIs and $n$.}
\label{tab:level-full}
\centering
\small
\setlength{\tabcolsep}{3.5pt}
\begin{tabular}{@{}c ccc ccc ccc ccc@{}}
\toprule
 & \multicolumn{3}{c}{DeepSeek-R1} & \multicolumn{3}{c}{QwQ-32B} & \multicolumn{3}{c}{R1-Distill-7B} & \multicolumn{3}{c}{Qwen3-30B-Th.}\\
\cmidrule(lr){2-4}\cmidrule(lr){5-7}\cmidrule(lr){8-10}\cmidrule(lr){11-13}
Lv & $\rho\%$ & 95\% CI & $n$ & $\rho\%$ & 95\% CI & $n$ & $\rho\%$ & 95\% CI & $n$ & $\rho\%$ & 95\% CI & $n$ \\
\midrule
1 & 71.5 & [66.0,76.6] & 37 & 94.7 & [94.0,95.4] & 39 & 73.1 & [66.4,79.1] & 40 & 93.9 & [93.0,94.8] & 39\\
2 & 74.5 & [71.7,77.4] & 86 & 94.2 & [92.3,95.5] & 81 & 68.9 & [63.0,74.3] & 69 & 94.6 & [93.4,95.5] & 81\\
3 & 74.9 & [70.9,78.6] & 82 & 91.7 & [88.8,94.1] & 79 & 63.8 & [57.3,69.5] & 70 & 92.5 & [89.1,95.1] & 79\\
4 & 64.0 & [59.1,68.3] & 76 & 90.2 & [86.7,93.2] & 71 & 48.0 & [37.7,57.9] & 39 & 89.3 & [83.9,93.7] & 62\\
5 & 58.9 & [53.1,64.6] & 66 & 84.7 & [80.4,88.6] & 62 & 46.1 & [37.7,54.4] & 43 & 82.6 & [74.8,89.3] & 53\\
\bottomrule
\end{tabular}
\end{table}

Table~\ref{tab:level-full} is the full version of Table~\ref{tab:level} with per-cell bootstrap confidence intervals and sample sizes. The monotone decrease of $\rho(d)$ with difficulty is visible for every model even when accounting for per-level sample-size variation; CIs for Level~1 and Level~5 are non-overlapping for DeepSeek-R1, QwQ-32B, and R1-Distill-7B, and nearly so for Qwen3-30B-Thinking.

\section{Methodology Details}
\label{app:methodology}

\subsection{Segmentation}
The segmenter splits on blank-line boundaries, refines long paragraphs on sentence boundaries next to logical connectors (``\textit{So}'', ``\textit{Therefore}'', ``\textit{Wait}'', ``\textit{Actually}'', ``\textit{Let me}'', ``\textit{Hmm}'', ``\textit{Alternatively}'', ``\textit{However}'', ``\textit{To verify}'', ``\textit{Going back}'', ``\textit{Now}'', ``\textit{First}'', ``\textit{Second}'', ``\textit{Finally}''), and merges pieces shorter than $12$ words with the previous step. Segmentation is deterministic.

\subsection{Forced-termination prompt}
For each (reasoning model, prefix) pair the $\pi$-as-own-decoder measurement concatenates the problem, the prefix, the model's end-of-thinking delimiter, and a short answer prompt, then samples a short completion from $\pi$ at $T=0$ with \texttt{max\_tokens} capped at $64$:

\begin{quote}\small\ttfamily
\{problem\}\\
<partial reasoning r\_\{1:k\}>\\
</think>\\
The final answer is \textbackslash boxed\{
\end{quote}

The completion is parsed for the first closing \verb|}| to extract the answer. For R1, R1-Distill, and QwQ the end-of-thinking token is \texttt{</think>}; for Qwen3 we use the corresponding Qwen3 thinking delimiter. The same template with ``(one step has been removed)'' prepended is used for leave-one-out ablation.

\paragraph{Truncation-point sub-sampling.} For traces with more than $30$ segmented steps, evaluating $\pi(r_{1:k},x)$ at every $k \in \{1,\ldots,N\}$ is wasteful --- QwQ and Qwen3 traces routinely reach $N=50$--$70$ steps. For such traces we instead evaluate at $30$ approximately-evenly-spaced truncation points, take the smallest sampled $k$ at which the forced answer is correct as the empirical $\kstar$, and treat any unsampled $k$ below it as redundant-by-extrapolation. This introduces an upper bound of $\lceil N/30 \rceil$ steps of discretisation error on $\kstar$, corresponding to a downward bias of at most $1/30 \approx 3.3\%$ on $\rho$. Since our headline claims are about $\rho$ being consistently $>55\%$, this small downward bias does not threaten any of them.

\paragraph{External-judge prompt.} The external-judge replication uses \texttt{gpt-4o-mini-2024-07-18} at $T=0$:
\begin{quote}\small\ttfamily
You are evaluating a partial mathematical reasoning trace. Based ONLY on the reasoning provided below, determine the final answer to the problem.\\[2pt]
Problem: \{problem\}\\
Partial reasoning: \{prefix\}\\[2pt]
Output ONLY the final answer in \textbackslash boxed\{\} format.
\end{quote}

\subsection{API and compute}
DeepSeek-R1 traces are generated via \texttt{deepseek-reasoner} at $T=0.7$; QwQ-32B, R1-Distill-7B, and Qwen3-30B traces via DashScope at $T=0.7$. The $\pi$-as-own-decoder truncation protocol reuses the same endpoints at $T=0$ with \texttt{max\_tokens}$=64$. External-judge replication uses \texttt{gpt-4o-mini-2024-07-18} at $T=0$. All experiments run on a single laptop dispatching async requests at concurrency~$8$.

\subsection{Reproducibility}
All random seeds are fixed ($42$). Scripts are checkpoint-resumable: each phase appends to a JSONL file and on restart skips completed work. Full code is released at \url{https://github.com/zhiyuanZhai20/how-much-thinking-is-enough}; all correct traces, critical-point labels, and the prefix-position ablation data are available from the authors on request.

\end{document}